\documentclass[letterpaper]{article} 
\usepackage{aaai25}  
\usepackage{times}  
\usepackage{helvet}  
\usepackage{courier}  
\usepackage[hyphens]{url}  
\usepackage{makecell}
\usepackage{booktabs}
\usepackage{graphicx} 
\usepackage{natbib}  
\usepackage{caption} 
\usepackage{algorithm}
\usepackage{algorithmic}
\usepackage{booktabs}
\usepackage{amsmath}
\usepackage{amssymb}
\usepackage{xcolor}
\usepackage{newfloat}
\usepackage{listings}
\DeclareCaptionStyle{ruled}{labelfont=normalfont,labelsep=colon,strut=off} 
\floatstyle{ruled}
\newfloat{listing}{tb}{lst}{}
\floatname{listing}{Listing}
\title{DreamFit: Garment-Centric Human Generation via a Lightweight Anything-Dressing Encoder}
\author{
    Ente Lin\textsuperscript{\rm 1}$^\star$, 
    Xujie Zhang\textsuperscript{\rm 2}$^\star$, 
    Fuwei Zhao\textsuperscript{\rm 3}, 
    Yuxuan Luo\textsuperscript{\rm 3}, 
    Xin Dong\textsuperscript{\rm 3}, \\
    Long Zeng\textsuperscript{\rm 1}$^\dagger$, 
    Xiaodan Liang\textsuperscript{\rm 2,4}$^\dagger$
}

\affiliations{
    \textsuperscript{\rm 1}Shenzhen International Graduate School, Tsinghua University \\
    \textsuperscript{\rm 2}Shenzhen Campus of Sun Yat-sen University \\
    \textsuperscript{\rm 3} ByteDance 
    \textsuperscript{\rm 4}Sun Yat-sen University \\

    linet22@mails.tsinghua.edu.cn,
    zhangxj59@mail2.sysu.edu.cn,\\
    zhaofuwei.777@bytedance.com, 
    luoyuxuan@bytedance.com, 
    dongxin.1016@bytedance.com, \\ 
    zenglong@sz.tsinghua.edu.cn, 
    xdliang328@gmail.com,
}

\usepackage{bibentry}

\begin{document}

\twocolumn[{
\renewcommand\twocolumn[1][]{#1}%
    \maketitle
    \centering
    \captionsetup{type=figure}
    \includegraphics[width=0.92\textwidth]
    {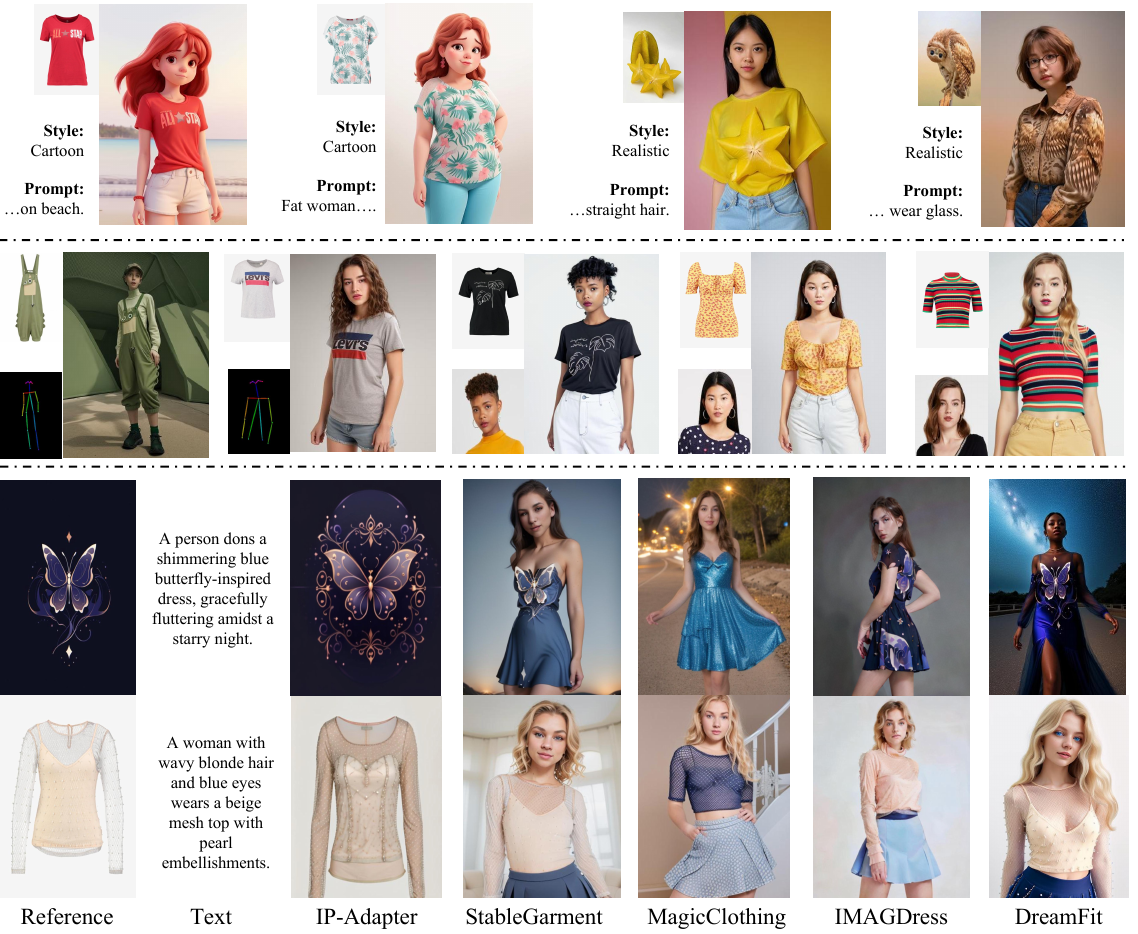}
    \captionof{figure}{Garment-centric human generation results of our DreamFit: \textbf{TOP:} DreamFit can synthesize human images with varied styles, backgrounds, and body shapes complying with the given clothing image and prompt. \textbf{Middle:} DreamFit is compatible with community plugins such as ControlNet \cite{zhang2023adding} and FaceID \cite{ye2023ip}. \textbf{Bottom:} DreamFit demonstrates superior performance compared to SOTA methods, achieving the highest levels of texture and texts consistency.}
    \label{fig:teaser}
}]
\renewcommand{\thefootnote}{}
\footnotetext{$^\star$ Equal contribution, interns at ByteDance.}
\footnotetext{$^\dagger$ Corresponding Authors.}

\begin{abstract}

Diffusion models for garment-centric human generation from text or image prompts have garnered emerging attention for their great application potential.
However, existing methods often face a dilemma: lightweight approaches, such as adapters, are prone to generate inconsistent textures; while finetune-based methods involve high training costs and struggle to maintain the generalization capabilities of pretrained diffusion models, limiting their performance across diverse scenarios. To address these challenges, we propose DreamFit, which incorporates a lightweight Anything-Dressing Encoder specifically tailored for the garment-centric human generation. 
DreamFit has three key advantages: (1) \textbf{Lightweight training}: with the proposed adaptive attention and LoRA modules, DreamFit significantly minimizes the model complexity to 83.4M trainable parameters. (2)\textbf{Anything-Dressing}: Our model generalizes surprisingly well to a wide range of (non-)garments, creative styles, and prompt instructions, consistently delivering high-quality results across diverse scenarios. (3) \textbf{Plug-and-play}: DreamFit is engineered for smooth integration with any community control plugins for diffusion models, ensuring easy compatibility and minimizing adoption barriers.
To further enhance generation quality, DreamFit leverages pretrained large multi-modal models (LMMs) to enrich the prompt with fine-grained garment descriptions, thereby reducing the prompt gap between training and inference. We conduct comprehensive experiments on both $768 \times 512$ high-resolution benchmarks and in-the-wild images. DreamFit surpasses all existing methods, highlighting its state-of-the-art capabilities of garment-centric human generation.

\end{abstract}

\section{Introduction}

Garment-centric human generation aims to synthesize high-quality images of stylized humans based on diverse combinations of text and image prompts, with a focus on accurately depicting the given garment details, as shown in Figure~\ref{fig:teaser}. 
This technology has gained significant traction in industries such as creative advertising and fashion design and sparks active exploration in the academic community \cite{wang2024stablegarment, chen2024magic}.

Currently, this task faces two critical challenges: the resource-intensive process of full model fine-tuning and the discrepancy between training and inference text prompts. 
Specifically, finetune-based methods like StableGarment\cite{wang2024stablegarment} rely on a full copy of diffusion UNet as the garment encoder, namely the ``ReferenceNet'', which consumes high GPU memory and becomes intractable when scaling up, e.g., from SD1.5\cite{SDInpainting} to SDXL\cite{podell2023sdxl}. Moreover, directly fine-tuning the entire UNet is parameter inefficient, requires more training time, and is prone to destroying the pretrained priors. As a result, the extracted garment feature is less descriptive, making it difficult to fully preserve the texture details (Figure~\ref{fig:teaser} Bottom). Alternatively, some lightweight methods \cite{ye2023ip, mou2024t2i} attempt to bypass the bulky ReferenceNet by directly mapping image features into the CLIP \cite{radford2021learning} latent space to control the generation through cross attention. Despite training faster, they tend to overlook crucial image feature details. Therefore, the trade-off between training efficiency and visual detail preservation is a significant problem for garment-centric human generation.

\begin{figure}
  \centering
  \includegraphics[width=1.0\hsize]{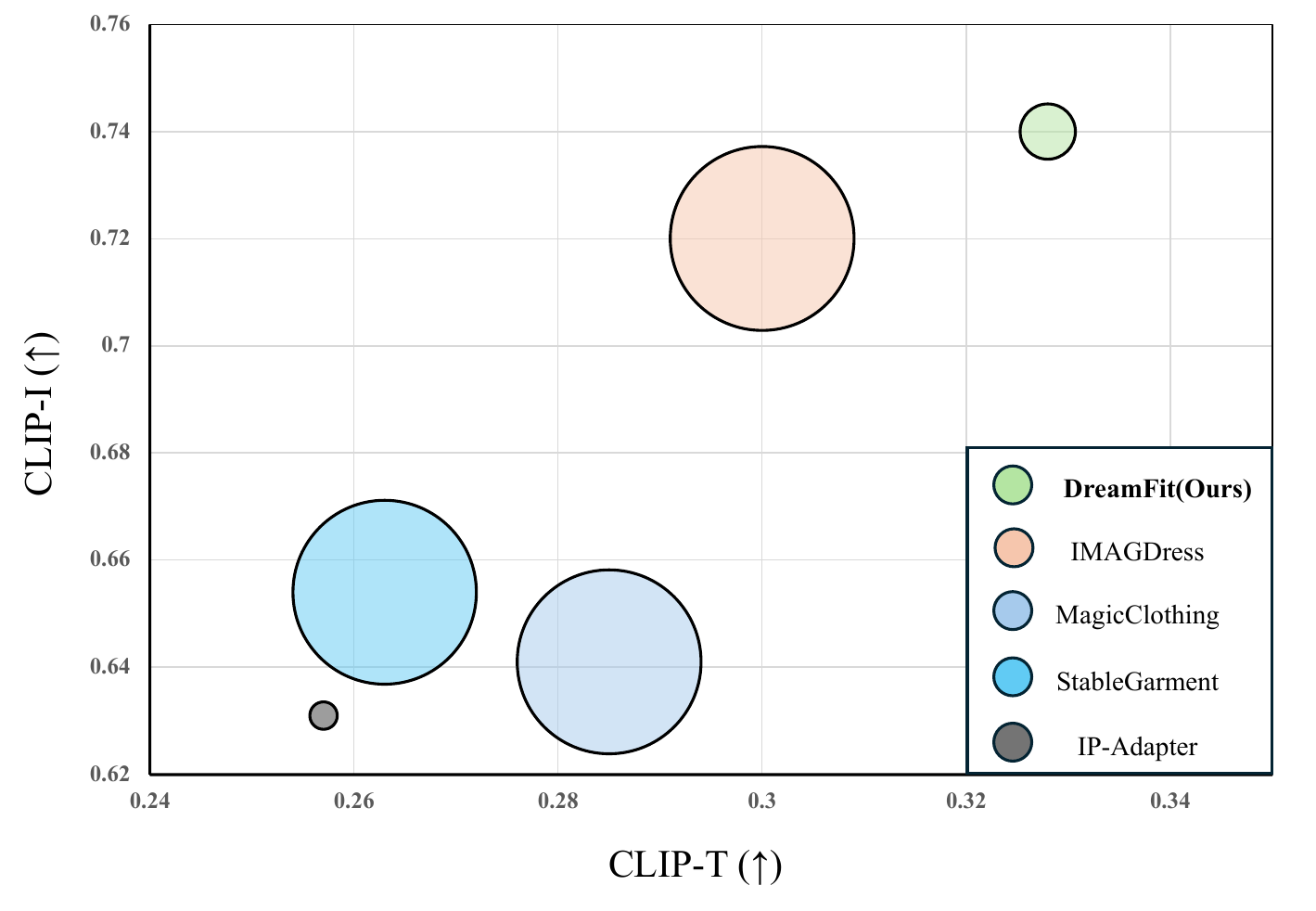}
  \vspace{-6mm}
  \caption{Performance comparison between baselines and our DreamFit. The circle size represents the number of trainable parameters, with larger circles indicating a higher parameter count. Higher CLIP-I and CLIP-T scores signify better alignment between the generated results and text descriptions. Our method not only achieves the best performance but also maintains much fewer training parameters.}
   \vspace{-4mm}
  \label{fig:motivation}
\end{figure}

Another challenge is the domain gap between text prompts for training and inference. To fully exploit the rich prior of pretrained T2I diffusion models, training prompts generated by Large Multimodal Models (LMMs) are typically comprehensive \cite{choi2024improving}. However, in inference, users often resort to simplified text due to the difficulty of manually creating such rich prompts. This gap further complicates the garment-centric generation process.


To address these challenges, we introduce DreamFit, a novel framework specifically designed for garment-centric human generation. 
DreamFit leverages a lightweight Anything-Dressing Encoder, which is derived through the activation of LoRA layers within the denoising UNet.
This design eliminates the need for resource-intensive components like ReferenceNet and effectively utilizes the rich priors of the diffusion model to extract fine-grained garment features. The extracted garment features are then integrated into the diffusion model through a novel adaptive attention mechanism, enabling high-quality and texture-consistent human image generation.
Furthermore, to bridge the gap between training and inference text prompts, DreamFit incorporates LMMs into the inference pipeline. This integration minimizes discrepancies, ensuring that the generated images maintain high fidelity and consistent quality across diverse scenarios.

As shown in Figure \ref{fig:motivation}, DreamFit beats all baselines at CLIP-I and CLIP-T metrics, showing the best text and texture consistency of generation results even with 10x fewer trainable parameters (83.4M v.s. 875M full model fine-tuning). Our main contributions are summarized as follows:

\begin{figure*}
  \centering
    \captionsetup{type=figure}
    \includegraphics[width=1.0\textwidth]{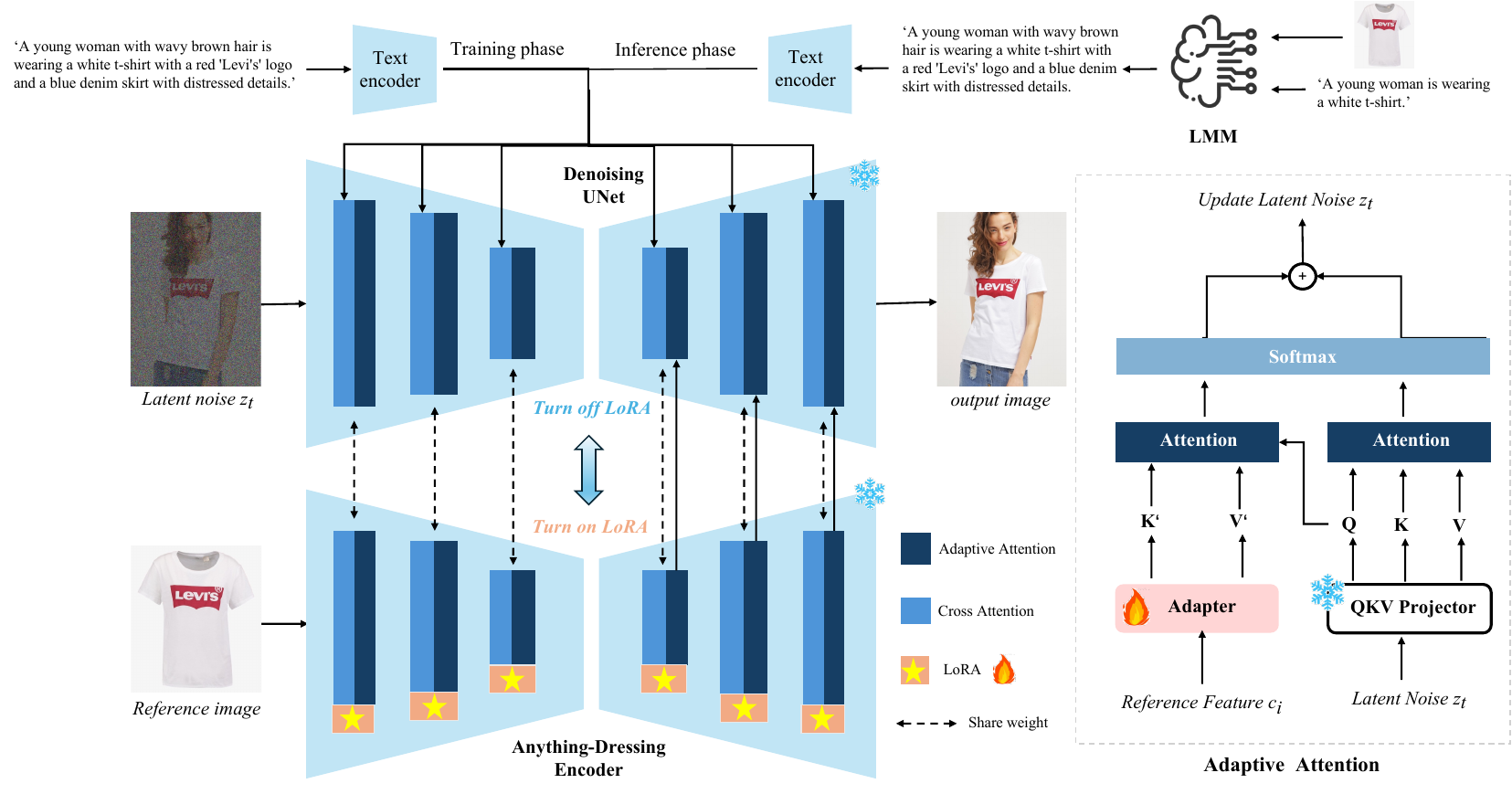}
  \caption{
Overview of DreamFit. Our method constructs an Anything-Dressing Encoder utilizing LoRA layers. The reference image features are extracted by the Anything-Dressing Encoder and then passed into the denoising UNet via adaptive attention.  Furthermore, we incorporate Large Multimodal Models (LMM) into the inference process to reduce the text prompt gap between the training and testing.
  } 
  \label{fig:architecture}
\vspace{-5mm}
\end{figure*}

\begin{itemize}
    \item We propose \textbf{DreamFit}, an efficient and dedicated framework for garment-centric human generation, enabled by a lightweight and plug-and-play Anything-Dressing Encoder. Our model addresses the inefficiencies and limitations of existing methods, particularly in aspects of model fine-tuning and computational efficiency.
    \item We integrate LMMs into the inference pipeline that effectively reduces the prompt gap between training and testing, enhancing the overall quality and fidelity of the generated results.
    \item Extensive experiments on open and internal benchmarks of $768 \times 512$ resolution verify the superiority of DreamFit, demonstrating state-of-the-art performance and robust generalization in diverse human generation tasks.
\end{itemize}

\section{Related work}

Human generation has witnessed rapid advancements, particularly with the emergence of controllable text-to-image generation technologies. Existing methods in this field often concentrate on either enhancing lightweight efficiency or enhancing texture consistency. However, these approaches struggle to balance the trade-offs between computational requirements and image quality.

Lightweight methods like \cite{hu2021lora, ye2023ip, mou2024t2i, zhang2023adding, ruiz2023dreambooth, chen2024anydoor,liu2024ada,han2024face,yang2024lora,li2024swiftdiffusion,xing2024tryon} have been proposed as lightweight approaches to achieve controllable human generation. These methods typically align image features with text features, using a combination of text and visual prompts to guide the generation process. T2I-adapter \cite{mou2024t2i} designed a series of convolutional layers to downsample and extract the reference image features, which are then fed into the denoising UNet via residual connections. IP-Adapter \cite{ye2023ip} introduces a sophisticated decoupled cross-attention mechanism to integrate image features into the denoising UNet. Although these methods excel in lightweight design, they often struggle to maintain high texture consistency, especially in intricate garment details. In contrast, our method not only retains the same level of lightweight efficiency but also ensures significantly improved texture consistency.

Additionally, several  methods \cite{zhu2023tryondiffusion,morelli2023ladi,zhang2023warpdiffusionefficientdiffusionmodel,wang2024mv,xu2024ootdiffusion,zhu2024m,kim2023stableviton,zhang2024mmtryon,hu2024animate,cui2023street} aiming for better texture consistency have chosen to utilize the UNet from Stable Diffusion as the image encoder to capture image prompt features. StableGarment \cite{wang2024stablegarment} finetunes a trainable copy of denoising UNet as the garment encoder and incorporates garment features using additive self-attention. MagicClothing \cite{chen2024magic} proposes joint classifier-free guidance to obtain a trade-off between garment features and text prompts. While these approaches, which require extensive finetuning of pretrained models such as those based on Stable Diffusion, yield commendable results in terms of quality, they come with significant drawbacks. These methods are resource-intensive, demanding significant memory resources and exhibiting sluggish training procedures, thus limiting their scalability for real-world applications.

Our work builds on these developments by leveraging frozen large diffusion models (LDMs) for garment-centric human image synthesis. We introduce a lightweight, plug-and-play Anything-Dressing Encoder that addresses the inefficiencies and limitations of existing methods, offering a scalable and flexible solution for garment-centric human image synthesis.

\section{Methodology}

\subsection{Preliminary}
\label{sec:SD}
Stable Diffusion is a text-conditioned latent diffusion model~\cite{rombach2022high}. Given a latent feature $z_0$ encoded from the input image by a VAE~\cite{kingma2013auto}, the forward diffusion process is firstly performed by adding noise according to a predefined noise scheduler $\alpha_t$~\cite{ho2020denoising}:
\begin{equation}
q(z_t|z_0) = \mathcal{N}(z_t; \sqrt{\alpha_t}z_0, (1 - \alpha_t)I).
\end{equation}
Then to reverse the diffusion process to generate new images, a noise estimator $\epsilon_\theta(\cdot)$ parameterized by an UNet is learned to predict the forward added noise $\epsilon$ with the objective function as follows:
\begin{equation}
\mathcal{L}_\text{dm} = \mathbb{E}_{\boldsymbol{z}_{0}, \boldsymbol{\epsilon}, \boldsymbol{c}_{t}, t}
\left[ ||\epsilon - \epsilon_\theta (z_t, t, c_t)||^2 \right],
\end{equation}
Where $c_t$ is the text condition associated with the image latent $z$. In the Stable Diffusion model, each block of the UNet consists of cross-attention and self-attention layers. The cross-attention layer facilitates attention between the image feature query and the text condition embedding, while self-attention layer operates within the image feature space.

\subsection{DreamFit}

The overall architecture of DreamFit is shown in Figure \ref{fig:architecture}. 
Our model consists of two main components: a frozen denoising UNet with pretrained stable diffusion weights, and a set of trainable LoRA layers. 
DreamFit takes as input a reference garment image and a text prompt that describes the detail of the output image.
The text prompt features, extracted by a frozen text encoder, are integrated into the denoising UNet via a cross-attention mechanism. The features of the reference image are extracted by a lightweight Anything-Dressing Encoder, which is obtained by turning on the LoRA layers in the denoising UNet. Once the LoRA layers are turned off, the lightweight Anything-Dressing Encoder reverts to the denoising UNet. The extracted reference features are then integrated into the denoising UNet through an adaptive self-attention mechanism. After multiple rounds of denoising, the denoising UNet generates an image that closely aligns with the reference image and text prompt. We elaborate the anything-dressing encoder and the adaptive attention in the following sections.

\textbf{Lightweight Anything-Dressing Encoder.}
The core concept of our model is that the pretrained diffusion model possesses extensive prior knowledge, making it a potential robust feature extractor. 
However, the diffusion model is designed to process latent noise, rendering them less effective for extracting features from noise-free reference images. 
To address this, we utilize LoRA layers to extend the feature extraction capability of the diffusion model to noiseless reference images.
Specifically, we incorporated the LoRA layers into the linear and convolutional layers of the denoising UNet, modifying the forward pass as follows:
\begin{equation}
    h = \phi (x) + I(x)* \Delta W (x)
\end{equation}

\begin{equation}
    I(x) = 
    \begin{cases} 
    0, & \text{if } x = z_t,\\
    1, & \text{if } x = c_i.
    \end{cases}
\end{equation}

Here, $\phi$ denotes the linear or convolutional layers in the denoising UNet.
$I(x)$ denotes a gate function and $\Delta W(x)$ denotes the added LoRA layer~\cite{hu2021lora}.
$z_t$ and $c_i$ represent the latent noise and the extracted reference features at each block of denoising UNet, respectively. 
With this approach, we only need to train the lightweight LoRA layers instead of a bulky ReferenceNet. And the denoising UNet is now seamlessly transformed into an Anything-Dressing Encoder with retained pretrained weights (i.e., better generalization capabilities).

\textbf{Adaptive attention.}
As shown in Figure \ref{fig:architecture}, the reference image features are integrated into the denoising UNet through adaptive attention. 
During training, the weights of our denoising UNet the frozen, which can hinder the pretrained attention module from effectively capturing the relationship between reference image features and the latent noise. 
Inspired by the adapter mechanism \cite{ye2023ip}, we introduced two trainable linear projection layers $\mathbf{W}_{k}^{\prime}$, $\mathbf{W}_{v}^{\prime}$, into the adaptive attention. These layers act as adapters to align the reference image features with latent noise. 
After applying cross-attention mechanisms, we add the cross-attention output to the self-attention output. The formulation of adaptive attention is:
\begin{equation}
\mathbf{z}_{t}^{\text {new }}=\operatorname{Softmax}\left(\frac{\mathbf{Q} \mathbf{K}^{\top}}{\sqrt{d}}\right) \mathbf{V}+\operatorname{Softmax}\left(\frac{\mathbf{Q}\left(\mathbf{K}^{\prime}\right)^{\top}}{\sqrt{d}}\right) \mathbf{V}^{\prime}
\end{equation}
where  
$\mathbf{Q}=\mathbf{W}_{q}\mathbf{z}_{t}, \mathbf{K}=\mathbf{W}_{k}\boldsymbol{z}_{t}, \mathbf{V}=\mathbf{W}_{v}\boldsymbol{z}_{t}, \mathbf{K}^{\prime}=\mathbf{W}_{k}^{\prime}\boldsymbol{c}_{i}, \mathbf{V}^{\prime}=\mathbf{W}_{v}^{\prime}\boldsymbol{c}_{i} $, and $\mathbf{W}_{q}, \mathbf{W}_{k}, \mathbf{W}_{v}$ are frozen linear projection layers in the denoising UNet. To accelerate the convergence of attention modules, we initialize the  $\mathbf{W}_{k}^{\prime}$, $\mathbf{W}_{v}^{\prime}$ with $\mathbf{W}_{k}$, $\mathbf{W}_{v}$.
This allows us to seamlessly and lightweightly inject the reference image features into the denoising UNet, resulting in synthesized images that are highly consistent with the reference image.

\begin{figure*}
  \centering
  \includegraphics[width=0.8\textwidth]{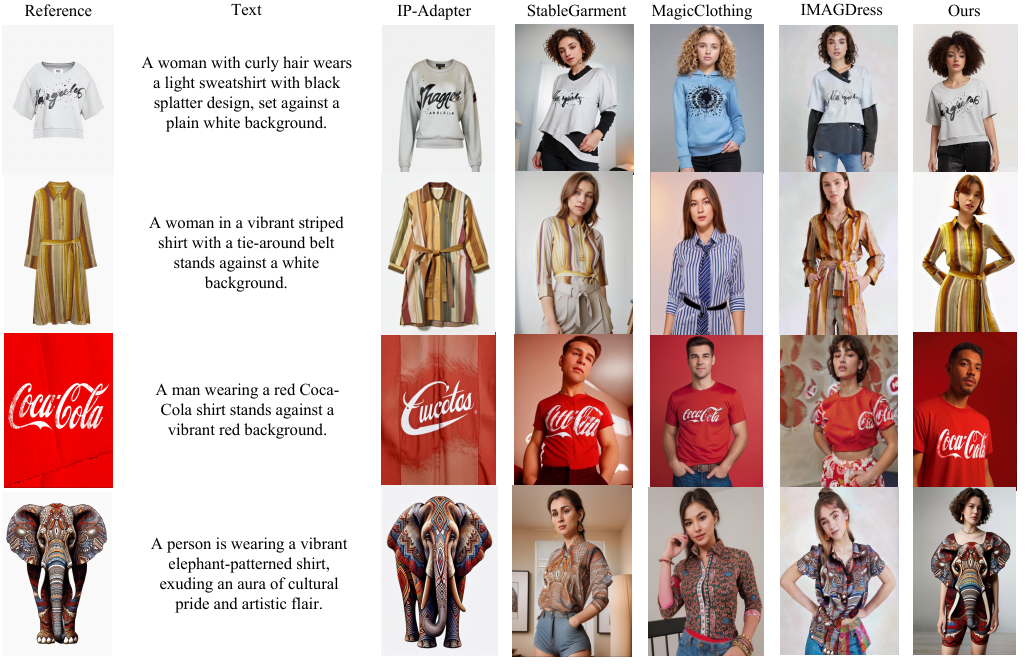}
  \vspace{-2mm}
  \caption{Qualitative comparison on the open and internal benchmarks. DreamFit demonstrates a distinct advantage in handling complex patterns and text. Please zoom in for more details.}
  \label{fig:comparsion}
\end{figure*}

\textbf{Training and inference}
During the training phase, we exclusively optimize the LoRA layers and the trainable adapters in the adaptive attention. Dreamfit is trained on a dataset containing image-text pairs, with the text generated by Language Model Models (LMMs), such as CogVLM \cite{wang2023cogvlm} and LLaVA1.5 \cite{liu2024improved}. To fully leverage the rich prior of the pretrained T2I diffusion model \cite{choi2024improving}, we prepare comprehensive captions for each image (e.g., "woman wears a long-sleeved polo shirt..."). The training objective remains consistent with the original diffusion model:
\begin{equation}
L_{\text {LDM }}=\mathbb{E}_{\boldsymbol{z}_{0}, \boldsymbol{\epsilon}, \boldsymbol{c}_{t}, \boldsymbol{c}_{i}, t}\left\|\boldsymbol{\epsilon}-\boldsymbol{\epsilon}_{\theta}\left(\boldsymbol{z}_{t}, \boldsymbol{c}_{t}, \boldsymbol{c}_{i}, t\right)\right\|^{2}
\end{equation}
During the inference phase, users are expected to input a reference image and a text description related to the desired output image. However, users often struggle to accurately describe fine-grained details in images and tend to provide simple text, leading to a domain gap between user input text and training text. This gap can result in decreased consistency and quality of the generated images. To address this challenge, we propose utilizing LMMs to rewrite the user input text based on the reference images. This approach enriches the text prompt with fine-grained descriptions of the garment, thereby reducing the prompt gap between training and inference. We denote the rewritten text as ${c}_{t}^{\prime}$. Then, we apply classifier-free guidance \cite{ho2022classifier} during the denoising process to generate the human image:
\begin{equation}
\hat{\boldsymbol{\epsilon}}_{\theta}\left(\boldsymbol{z}_{t}, \boldsymbol{c}_{t}^{\prime}, \boldsymbol{c}_{i}, t\right)=w \boldsymbol{\epsilon}_{\theta}\left(\boldsymbol{z}_{t}, \boldsymbol{c}_{t}^{\prime}, \boldsymbol{c}_{i}, t\right)+(1-w) \boldsymbol{\epsilon}_{\theta}\left(\boldsymbol{z}_{t}, t\right)
\end{equation}

\section{Experiment}

\subsection{Datasets}
To train Dreamfit, we collected approximately 500,000 garment-person image pairs from the internet and captioned them using large multi-modal models.
For model evaluation, we introduce two garment-centric human generation benchmarks derived from public datasets and the Internet. The open benchmark is constructed using a subset of VITON-HD \cite{choi2021viton} and DressCode \cite{morelli2022dress} test sets. Specifically, we handpicked 200 diverse garments from these datasets encompassing various styles, colors, shapes, and textures. Each garment is associated with refined prompts generated through large multi-modal models. 
In addition to the open benchmark, we further developed a more challenging internal benchmark. To build the internal benchmark, we manually gathered 200 reference images from the internet, including garments with intricate textures and backgrounds, and non-garment images featuring animals, fruits, and patterns. This benchmark is designed to evaluate the robustness and generalization capability of the model across diverse reference image types.


\begin{figure*}
  \centering
  \includegraphics[width=0.9\textwidth]{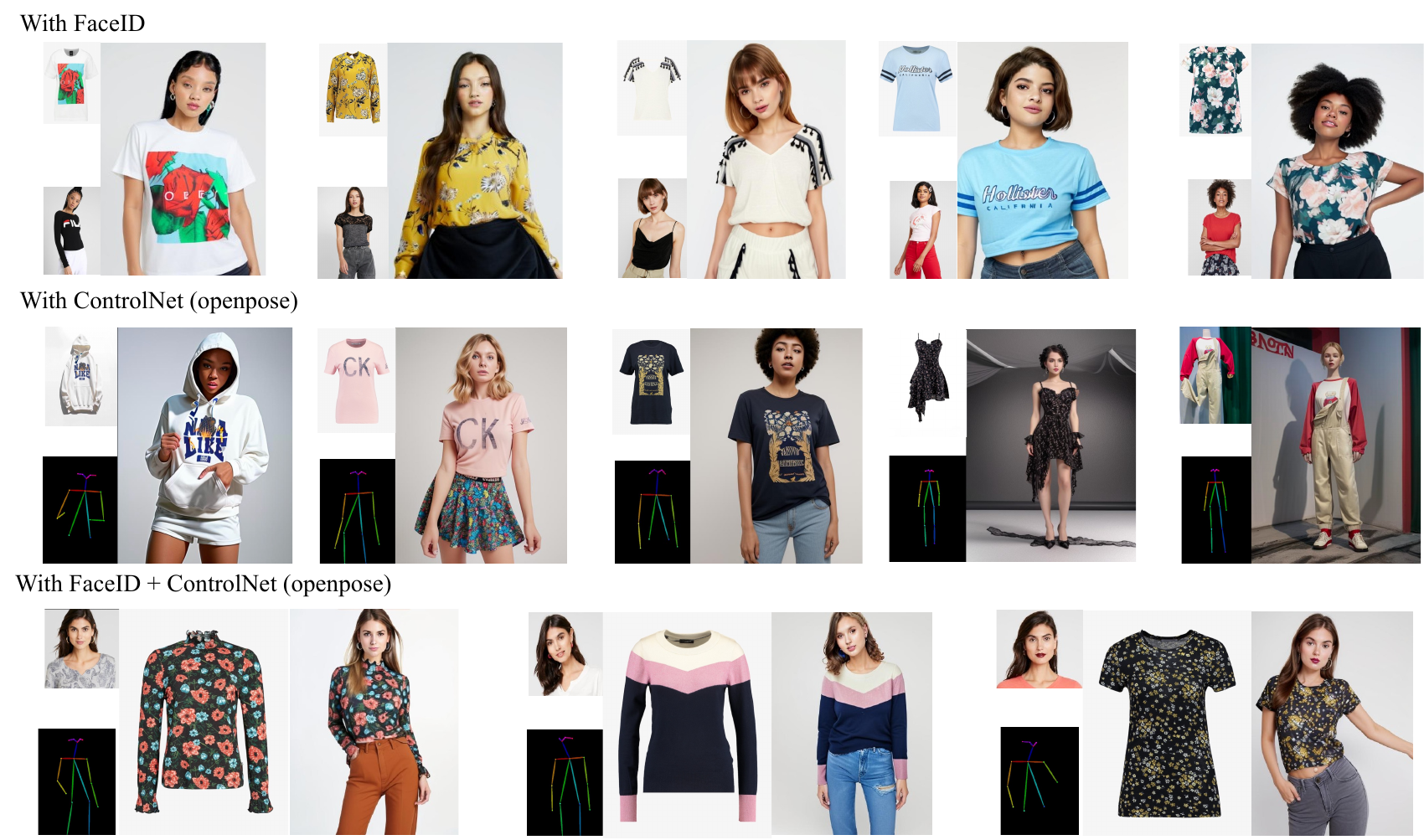}
  \vspace{-2mm}
  \caption{Plug-and-play results of DreamFit, our method can seamlessly integrate with community conditional control plugins.}
  \label{fig:more_result}
\end{figure*}

\subsection{Implementation Details}

The denoising UNet is initialized with the weights of SD1.5 and we use CLIP ViT-L/14 \cite{radford2021learning} as the text encoder. Our model was trained on paired images with a resolution of $768 \times 512$. We initialized the LoRA layers in the same manner as described in \cite{hu2021lora}, with the LoRA rank set to 64. The training was conducted on 8 A800 (40G) GPUs for 90k steps, with a batch size of 4 per GPU. We utilized the AdamW optimizer with a fixed learning rate of 1e-4. 
During inference, we use CogVLM \cite{wang2023cogvlm} to refine the user input text. We use DDIM \cite{song2020denoising} sampler with 50 steps and set guidance scale $w$ to 7.5. 
To validate scalability, we also initialized the Denoising UNet as SDXL and FLUX~\cite{flux2024}, more implementation details and results of SDXL and FLUX versions please refer to supplementary material.

\subsection{Baselines}
We selected four state-of-the-art methods, including StableGarment \cite{wang2024stablegarment}, MagicClothing \cite{chen2024magic}, IMAGDress \cite{shen2024imagdressing}, and IP-Adapter \cite{ye2023ip}. 
We directly use their released pretrained models for comparison. All experiments are conducted with a resolution of $768 \times 512$. 
For a fair comparison, we used the model initialized with the weights of SD1.5 in the experiment.

\subsection{Qualitative Results}
The qualitative comparison is illustrated in Figure~\ref{fig:comparsion}. The results demonstrate the superiority of our DreamFit over other baselines. Firstly, our method exhibits better texture and text consistency. Existing lightweight approaches, such as IP-Adapter, lose text and texture consistency when faced with reference images containing complex details. While finetune-based methods like StableGarment, MagicClothing, and IMAGDress surpass IP-Adapter significantly, they still struggle to generate human images with fine-grained features that highly match the text prompts.
Secondly, images generated by baselines generally exhibit issues such as blurriness (IMAGDress) and incorrect body proportions (third row of StableGarment column). In contrast, our method achieves the best quality and realism. To show the model's excellent compatibility, we provide additional results showcasing DreamFit integrated with community plugins in Figure~\ref{fig:more_result}.

\subsection{Quantitative Results}
\noindent\textbf{Metric.} 
We quantitatively measure the generated results from three aspects: consistency, generation quality, and human preference.
Regarding consistency, we employ CLIP-T and CLIP-I \cite{hessel2021clipscore} to evaluate text and texture consistency, respectively. We utilize Aesthetic score(AS)  \cite{schuhmann2022laion} to evaluate generation quality. Several studies~\cite{ku2023viescore, peng2024dreambench++} have shown that GPT can produce evaluation results that are remarkably consistent with humans. 
Consequently, we employ GPT-4o~\cite{achiam2023gpt} as the automatic evaluation of human preference, denoted as Human-Aligned Score (HAS). For more details on the evaluation, please refer to the supplementary material.
 

\noindent\textbf{Evaluation.} We conducted experiments using both open and internal benchmarks to assess the performance of our model against baselines. The quantitative results are reported in Tab~\ref{quantitative-results-table}. 
Our DreamFit consistently outperforms all baselines across various metrics and benchmarks, showcasing its ability to produce results with superior text and texture consistency, exceptional generation quality, and strong alignment with human preferences. 
Furthermore, compared with fully finetune-based methods, like  StableGarment, IMAGDress, DreamFit has the fewest trainable parameters while achieving the best performance. 
The quantitative result underscores DreamFit's resource efficiency and robustness.


\begin{table*}[ht]
    \centering
    \begin{tabular}{c|c|cccc|cccc}
        \toprule
        {Method} & {Trainable Parameters} & \multicolumn{4}{c|}{Open Benchmark} & \multicolumn{4}{c}{Internal Benchmark} \\ 
        \cmidrule(lr){3-6} \cmidrule(lr){7-10}
        & & CLIP-T$\uparrow$ & CLIP-I$\uparrow$ & AS$\uparrow$ & HAS$\uparrow$ & CLIP-T$\uparrow$ & CLIP-I$\uparrow$ & AS$\uparrow$ & HAS$\uparrow$ \\ 
        \midrule
        IP-Adapter & \textbf{22M} & 0.257 & 0.631 & 4.667 & 0.568 & 0.135 & 0.422 & 2.634 & 0.430 \\ 
        StableGarment & 859M & 0.263 & 0.654 & 4.877 & 0.768 & 0.275 & 0.629 & 5.238 & 0.683 \\ 
        MagicClothing & 875M & 0.285 & 0.641 & 5.007 & 0.443 & 0.320 & 0.583 & 5.462 & 0.595 \\ 
        IMAGDress & 875M & 0.300 & 0.720 & 4.867 & 0.783 & 0.290 & 0.632 & 5.204 & 0.840 \\ 
        Ours & 83.4M & \textbf{0.328} & \textbf{0.740} & \textbf{5.414} & \textbf{0.856} & \textbf{0.334} & \textbf{0.687} & \textbf{5.516} & \textbf{0.920} \\ 
        \bottomrule
    \end{tabular}
    \vspace{-3mm}
    \caption{Quantitative comparison of different methods on three metrics for both Open and Internal benchmarks.}
    \label{quantitative-results-table}
\end{table*}

\subsection{Ablation Study}

\begin{figure}
  \centering
  \includegraphics[width=1.0\hsize]{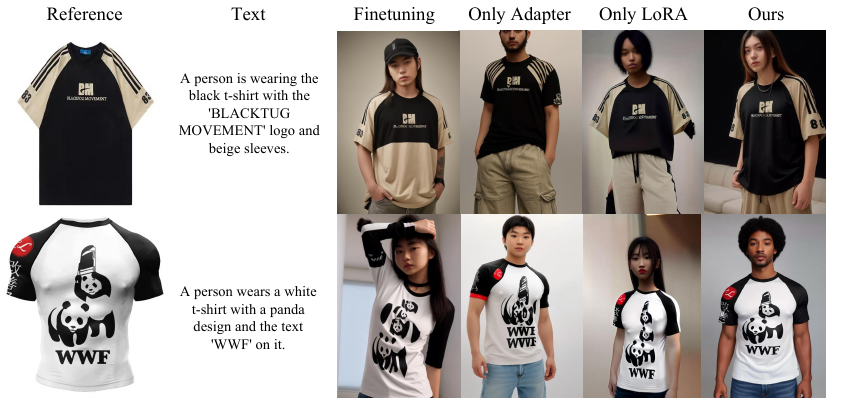}
  \caption{Qualitative results of the ablation study on network components.}
  \label{fig:ablation_components}
\end{figure}

\subsubsection{Effectiveness of Network Components}

To evaluate the impact of different network components on the final performance, we conducted an ablation study comparing various model configurations. We experiment with three setups: finetuning the entire Anything-Dressing Encoder (finetuning), only optimizing parameters in LoRA layers (only LoRA), and only optimizing parameters of adapters in adaptive attentions (only Adapter). The results are summarized in Tab~\ref{table:ablation-study1} and the qualitative results are shown in Figure~\ref{fig:ablation_components}.

Despite having the highest number of trainable parameters, finetuning the entire Anything-Dressing Encoder exhibited subpar performance on text and texture consistency. We conjecture this is because finetuning can compromise the model's pre-trained prior, leading to a decline of its feature extraction capability. While the "only LoRA" and "only Adapter" configurations showed better texture consistency than finetuning, they performed inadequately for generation quality. In contrast, our DreamFit, which combines and optimizes both LoRA and Adapter, achieves superior results across all metrics.

\begin{figure}
  \centering
  \includegraphics[width=1.0\hsize]{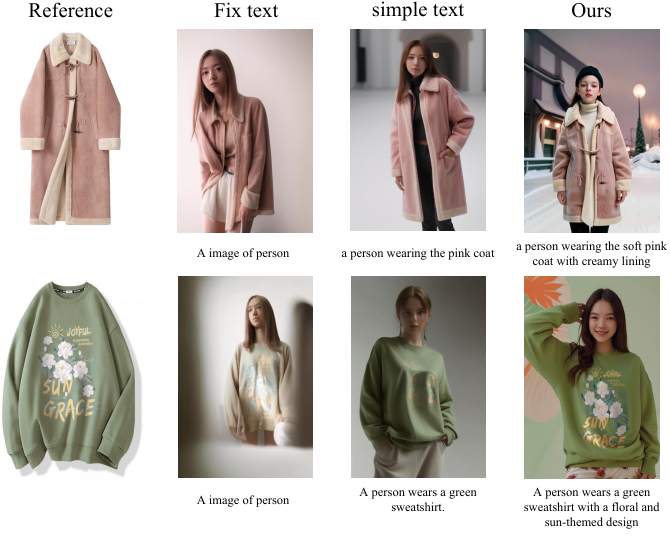}
  \caption{Qualitative results of the ablation study on text prompts.}
  \label{fig:ablation_text}
\end{figure}

\begin{table}[t]
\small
\centering
\setlength{\tabcolsep}{3pt} 
\renewcommand{\arraystretch}{0.9} 
\begin{tabular}{l c c c c}
  \toprule
  {Method} & \makecell{Trainable \\Parameters} & {CLIP-T$\uparrow$} & {CLIP-I$\uparrow$} & {AS$\uparrow$} \\
  \midrule
  Finetuning        & 875M  & 0.322 & 0.622 & \textbf{5.534} \\
  Only LoRA              & 66M   & 0.322 & 0.686 & 5.250 \\
  Only Adapter           & 17.4M & 0.310 & 0.679 & 5.230 \\
  Ours                   & 83.4M & \textbf{0.334} & \textbf{0.687} & 5.516 \\
  \bottomrule
\end{tabular}
\caption{Quantitative results of the ablation study on network
components}
\vspace{-5mm}
\label{table:ablation-study1}
\end{table}

\begin{table}[t]
\small
\centering
\begin{tabular}{l c c c}
  \toprule
  {Method} & {CLIP-I $\uparrow$} & {AS$\uparrow$} & {Human Study$\uparrow$} \\
  \midrule
  Fixed text       & 0.588 & 5.283 & 0.03 \\
  Simple text     & 0.663 & 5.186 & 0.10 \\
   Ours        & \textbf{0.687} & \textbf{5.516} & \textbf{0.87} \\
  \bottomrule
\end{tabular}
\caption{Quantitative comparison of different text prompt configurations.}
\label{table:ablation-study2}
\vspace{-5mm}
\end{table}

\subsubsection{Impact of Text Prompts}

To further explore the influence of text prompts, we conducted experiments with three different types of text inputs: (1) a fixed text prompt (Fix text), (2) a simple text prompt (Simple text) for simulating user input, and (3) text prompts rewritten by LMMs (Ours). The qualitative and quantitative results for each configuration are shown in Figure~\ref{fig:ablation_text} and Tab~\ref{table:ablation-study2}.

The results demonstrate that text prompts rewritten by large multi-modal models yield superior quality outputs across all metrics, including notably higher scores in human study evaluations. This underscores the efficacy of incorporating advanced LMMs to enhance prompts with detailed descriptions of reference images, thereby narrowing the gap between training and inference prompts and boosting the quality and realism of the generated images.

\section{Conclusion}

In this paper, we introduced \textbf{DreamFit}, a novel garment-centric human image generation framework designed to address the inefficiencies and limitations of existing methods. By leveraging a lightweight, plug-and-play Anything-Dressing Encoder based on LoRA layers, DreamFit significantly streamlines model complexity and memory usage, facilitating more efficient and scalable training procedures. Our approach integrates large multi-modal models into the inference process, effectively reducing the domain gap between training and inference text prompts and enhancing the overall quality and consistency of the generated images.
Extensive experiments conducted on open and internal benchmarks demonstrate that DreamFit not only achieves state-of-the-art performance but also exhibits superior generalization capabilities across diverse scenarios.

\section{Acknowledgments}

This work was sponsored by National Key Research and Development Program of China (Grant No.  2022YFB3303101), in part by Shenzhen Science and Technology Program (Grant No. GJHZ20220913142600001) and Doubao Fund.

\bibliography{aaai25}

\clearpage

\section{Supplementary Material}
\vspace{5mm}

\subsection{Implementation details}
In this section, we demonstrate how our framework is adapted to the SDXL and FLUX architectures.
\subsubsection{SDXL}
We initialized the Denoising UNet as SDXL (SDXL version) and trained the model on 4 A100 (80G) GPUs for 90k steps with a batch size of 8. The LoRA rank is set to 64. The AdamW optimizer with a constant learning rate of 1e-4 is used. The resolution of training image pairs is $1024 \times 768$. 

\subsubsection{FLUX}
We initialized the Denoising UNet as FLUX.dev \cite{flux2024} (FLUX version). We train the model on paired images with a resolution of $768 \times 512$. The LoRA rank is set to 32. The training was conducted on 8 A100 (80G) GPUs for 20k steps, with  DeepSpeed~\cite{deepspeed} ZeRO-2 to reduce memory usage, at a batch size of 4 per GPU. We also utilized the AdamW optimizer with a fixed learning rate of 1e-4. 

\subsection{Additional Results}\label{sec:AD}
\subsubsection{SD1.5}
To further demonstrate DreamFit's capabilities, we illustrate additional results on our Internal benchmark in Figure~\ref{fig:more_result_internal}. These results confirm DreamFit's impressive generalization ability across diverse reference images. In addition, we also provide more stylized generation results in Figure~\ref{fig:more_result_stylized}.

\subsubsection{SDXL and FLUX}\label{subsec:AD_SDXL_FLUX}
To further illustrate DreamFit's scalability, we show additional results for SDXL and FLUX on the open benchmark, as depicted in Figure~\ref{fig:more_result_SDXL} and Figure~\ref{fig:more_result_FLUX}. Compared to the SD1.5 version, the SDXL and FLUX versions exhibit superior generation quality. Moreover, we compared the performance of the SDXL and FLUX versions, as shown in Figure~\ref{fig:more_result_compare}. The qualitative results demonstrate that the FLUX version significantly surpasses SDXL in preserving fine-grained textures and logos. We believe this is due to FLUX's more powerful VAE and prior knowledge.

\subsection{Evaluation detail}\label{sec:Human_ed}
\subsubsection{Human-Aligned evaluation detail}\label{sec:HASED}
We employ GPT-4o-2024-05-13~\cite{achiam2023gpt} for Human-Aligned Evaluation (HAE).  The evaluation pipeline is illustrated in Figure~\ref{fig:HAS_pipeline}. 
To design the evaluation prompts, we drew inspiration from DreamBench++~\cite{peng2024dreambench++}, making adjustments to the task definitions, scoring criteria, and scoring range to better suit the Garment-Centric Human Generation task.
Figure~\ref{fig:HAS_visulize} showcases output examples along with corresponding evaluation rationales for different scores. 
GPT-4o assigns scores on a scale from 0 to 4, which are then normalized to a range of 0 to 1 for quantitative evaluation purposes.

\subsubsection{Human evaluation detail}\label{sec:Human_ed}
For human evaluation, we created a questionnaire for the garment-centric human generation task, as illustrated in  Figure~\ref{fig:human_sd}. We invited 100 volunteers to complete the questionnaire, which included 40  assignments. Specifically, Each assignment presented a reference image, and volunteers were asked to chose the image with the highest texture consistency and realism from three options generated by different text prompt configurations: (1) a fixed text prompt,
(2) a simple text prompt, and (3) text prompts rewritten by LMMs. The order of the generated results in each assignment was randomly shuffled.

\subsection{Societal impact}\label{sec:SI}
As the most generative approach, DreamFit can potentially be misused to create images that infringe on copyrights or privacy. Therefore, it's important to advocate for the responsible use of the model.

\subsection{Limitations and future work}\label{sec:L&FW}
While our method demonstrates strong resource efficiency, it still incurs OOM (Out Of Memory) when we train the FLUX version at 
higher resolutions such as $1024 \times 768$, even with 8 A100 (80G) GPUs.
In future work, we will concentrate on developing a more lightweight and high-quality method for garment-centric human generation. Additionally, we will apply our method to a wider range of conditional generation tasks, such as identity-preserving generation.



%


\setcounter{figure}{7}

\begin{figure*}
  \centering
  \includegraphics[width=0.9\textwidth]{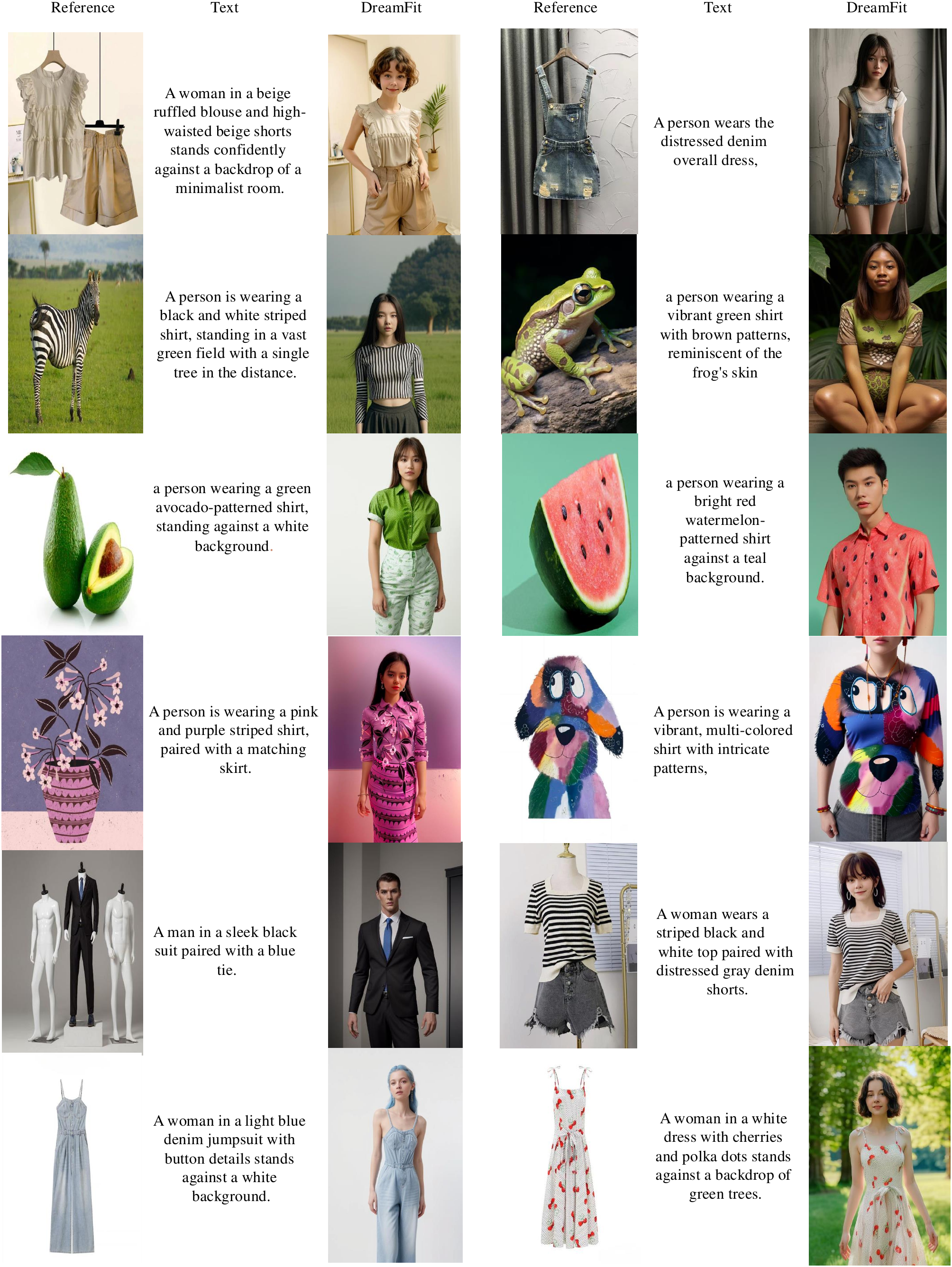}
  \caption{More qualitative results at 768 resolution on internal benchmark.}
  \label{fig:more_result_internal}
\end{figure*}

\begin{figure*}
  \centering
  \includegraphics[width=0.9\textwidth]{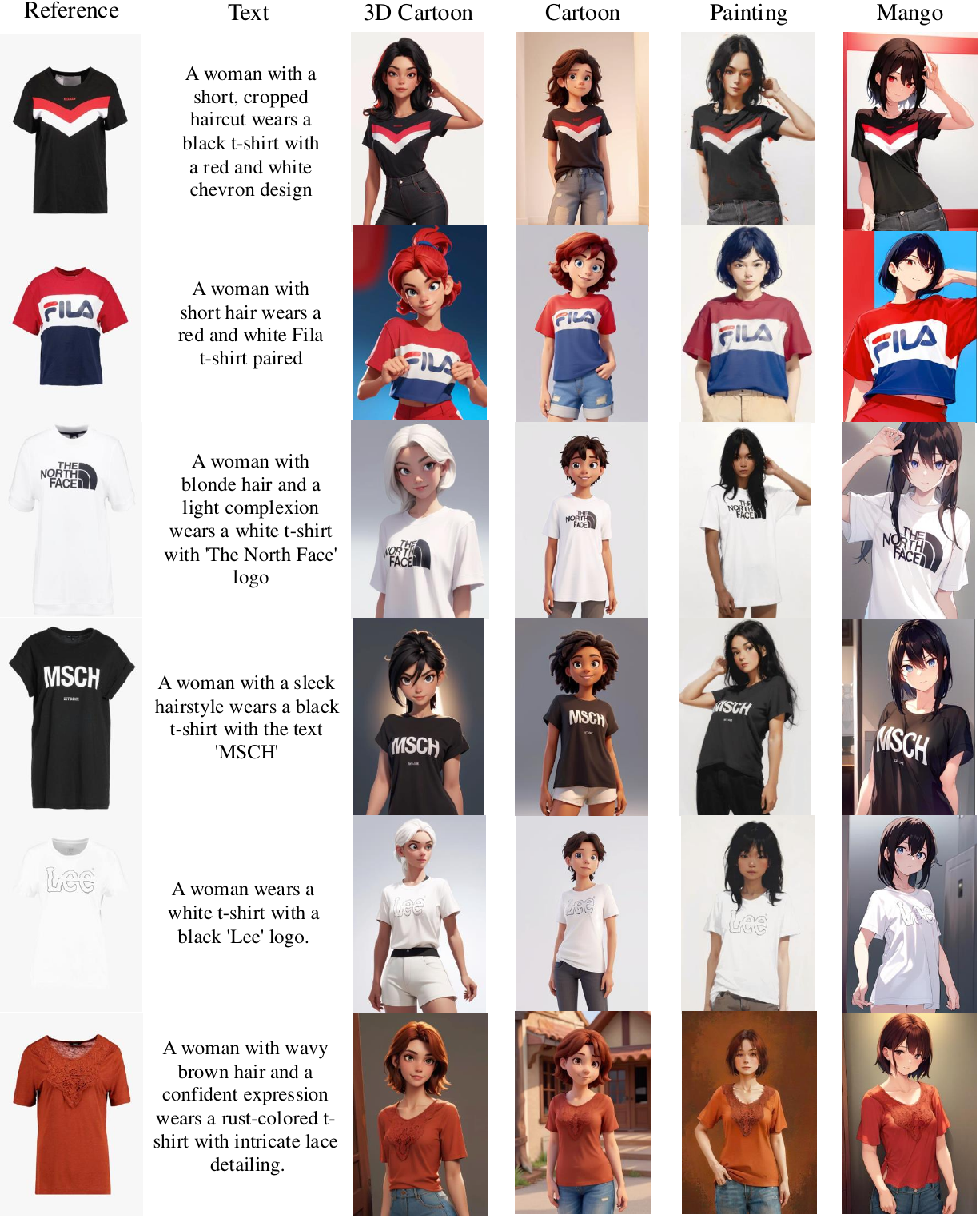}
  \caption{Qualitative results of SDXL version at 1024 resolution on the Open benchmark. Please zoom in for more details.}
  \label{fig:more_result_stylized}
\end{figure*}

\begin{figure*}
  \centering
  \includegraphics[width=0.9\textwidth]{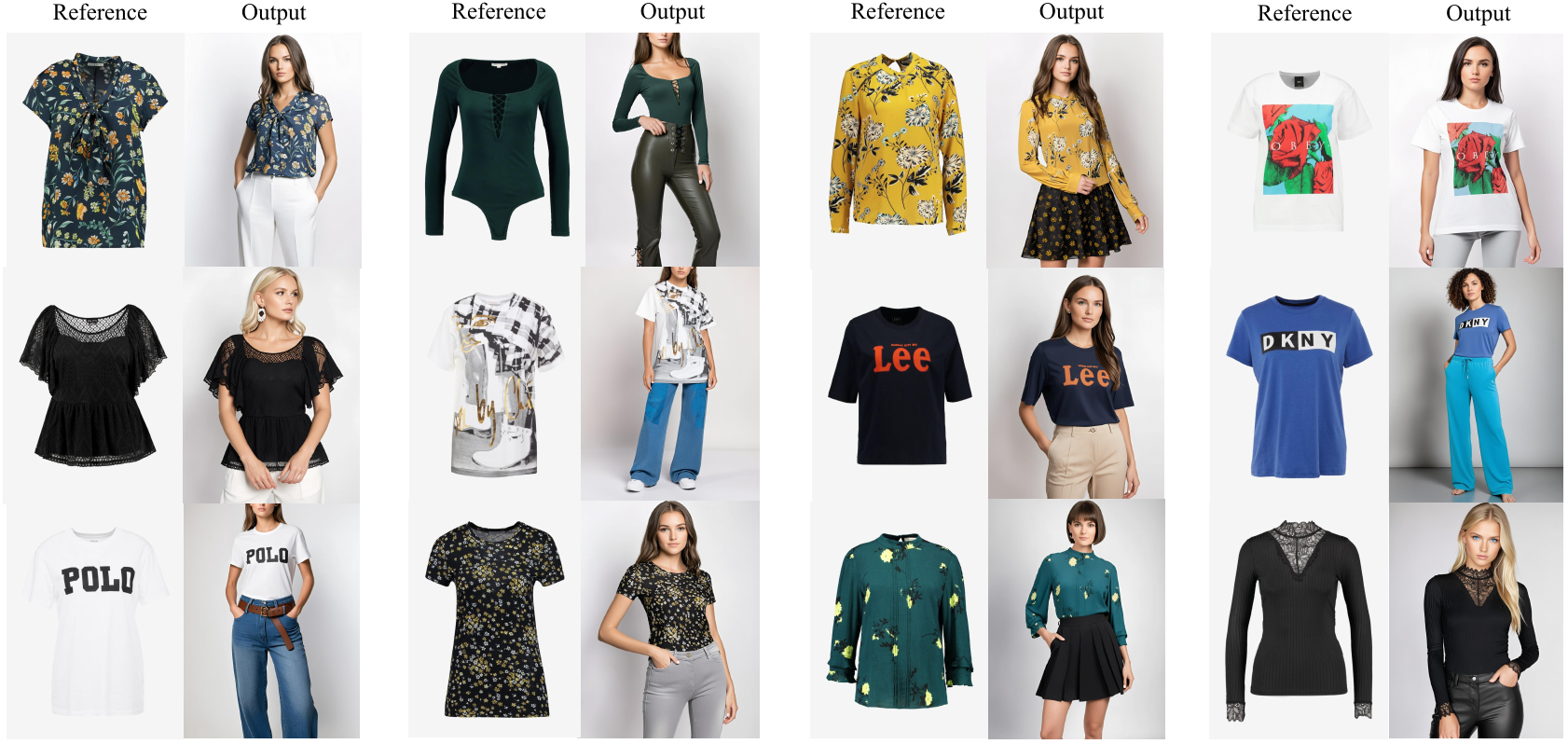}
  \caption{Qualitative results of SDXL version at 1024 resolution on the open benchmark. Please zoom in for more details.}
  \label{fig:more_result_SDXL}
\end{figure*}

\begin{figure*}
  \centering
  \includegraphics[width=0.9\textwidth]{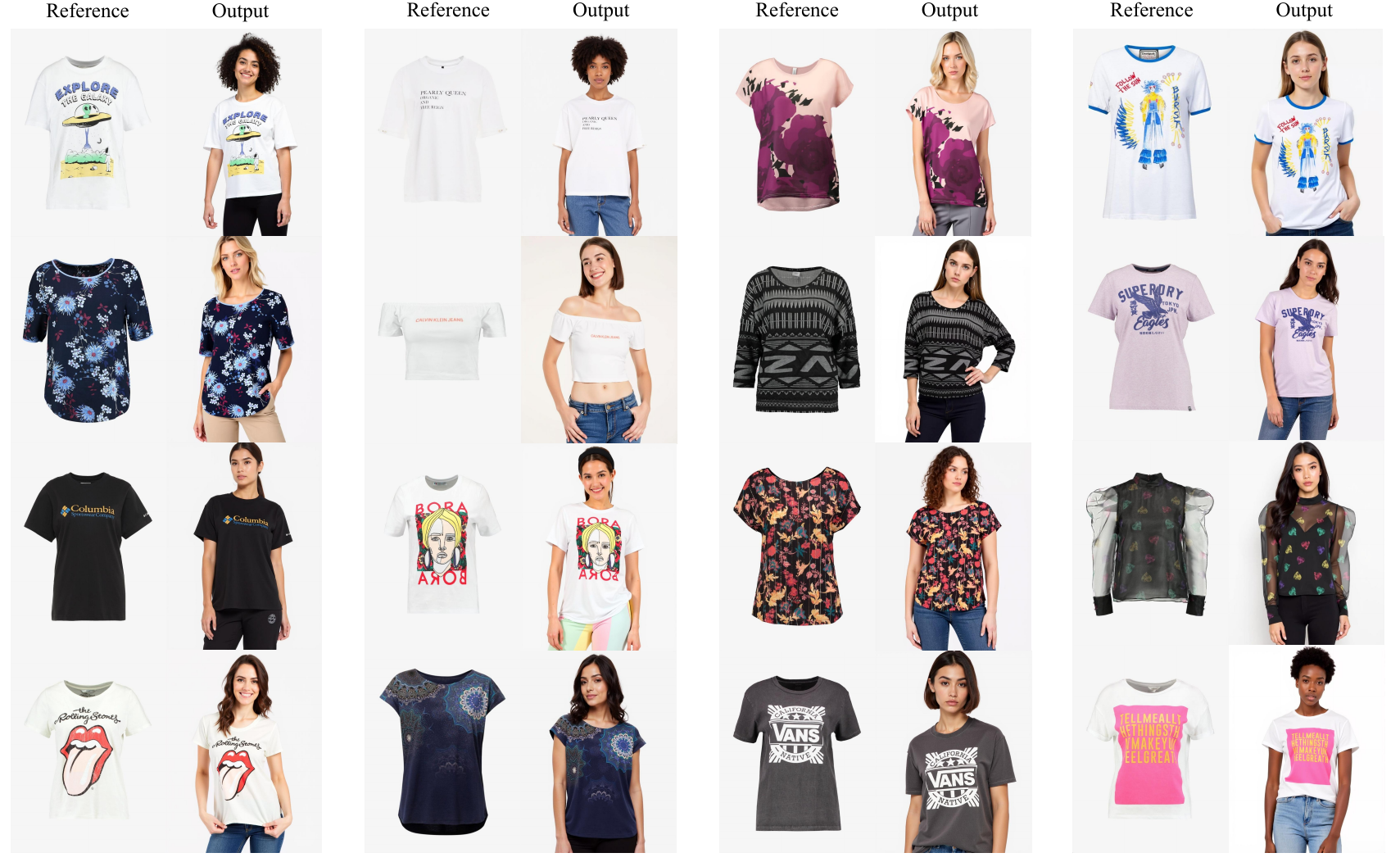}
  \caption{Qualitative results of FLUX version at 1024 resolution on the open benchmark. Please zoom in for more details.}
  \label{fig:more_result_FLUX}
\end{figure*}

\begin{figure*}
  \centering
  \includegraphics[width=0.9\textwidth]{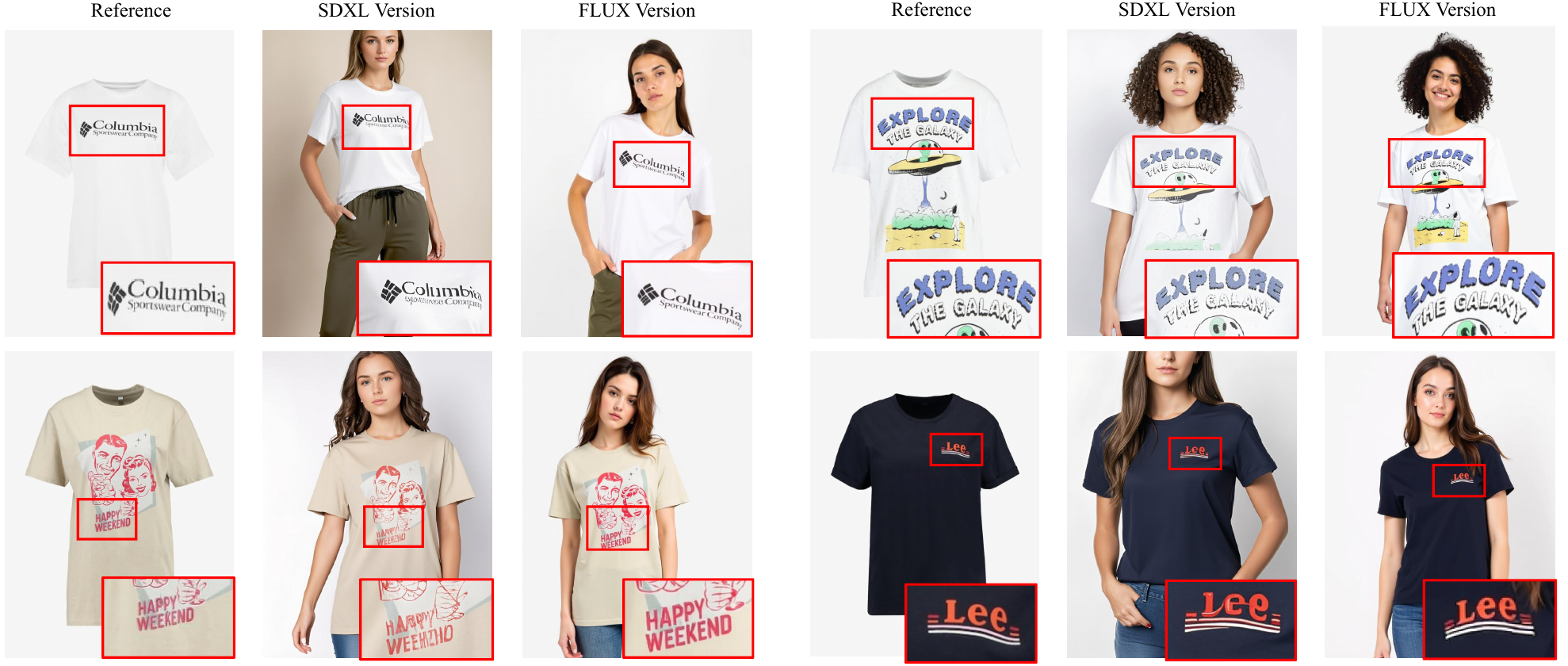}
  \caption{Qualitative comparison results between SDXL and FLUX versions at 1024 resolution. FLUX version significantly outperforms SDXL in maintaining fine-grained textures and logos. Please zoom in for more details.}
  \label{fig:more_result_compare}
\end{figure*}

\begin{figure*}
  \centering
  \includegraphics[width=0.9\textwidth]{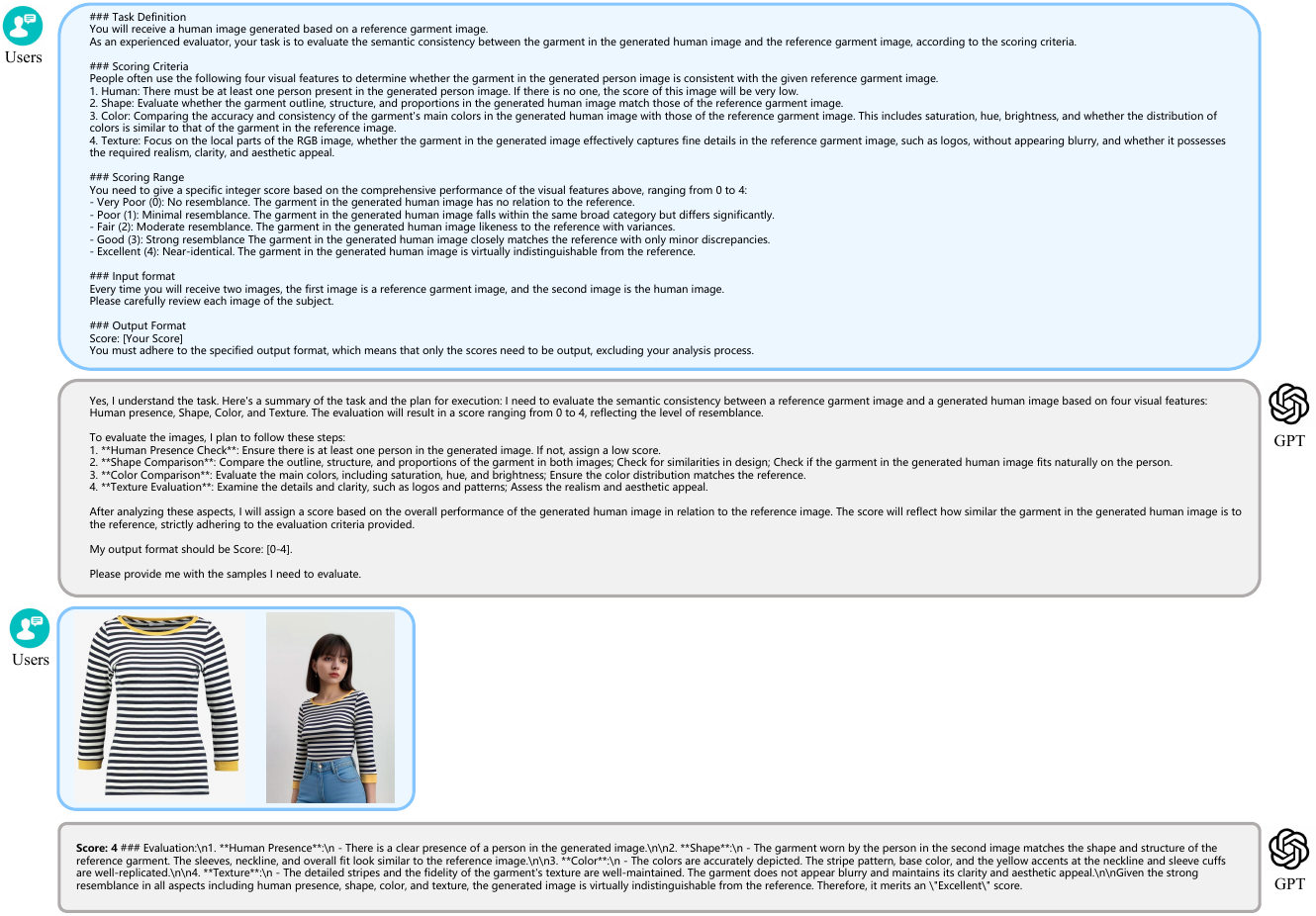}
  \caption{The Human-Aligned evaluation pipeline. We utilize GPT-4o for conducting automated evaluations.}
  \label{fig:HAS_pipeline}
\end{figure*}

\begin{figure*}
  \centering
  \includegraphics[width=0.6\textwidth]{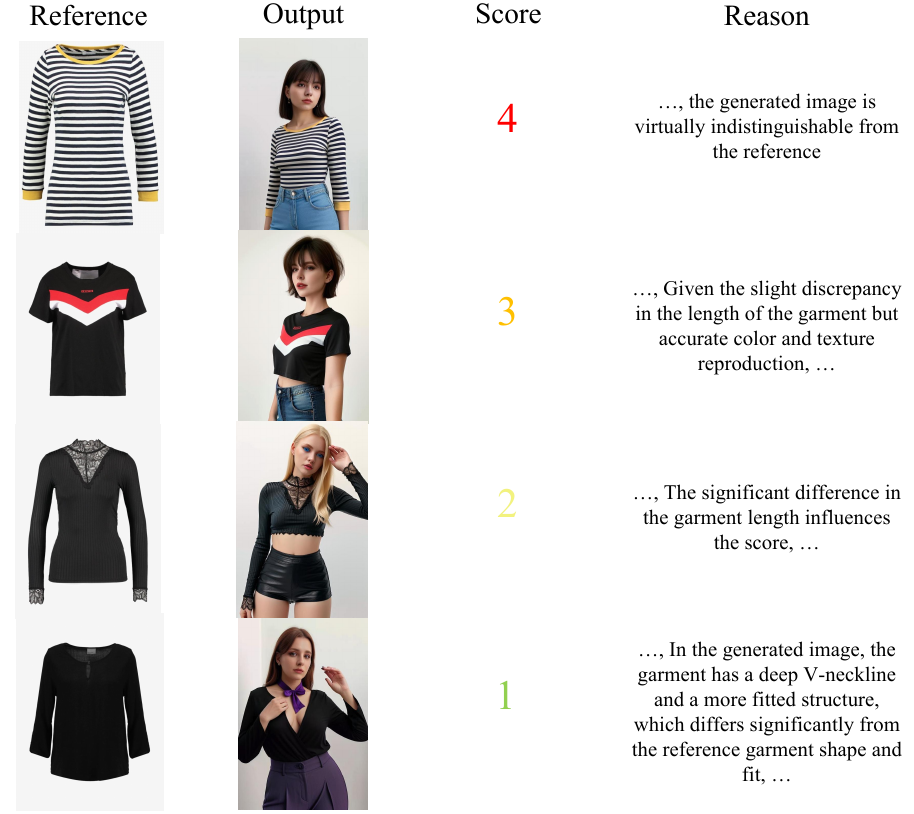}
  \caption{Visualized results of the outputs and evaluation reasons corresponding to different scores.}
  \label{fig:HAS_visulize}
\end{figure*}

\begin{figure*}
  \centering
  \includegraphics[width=0.7\textwidth]{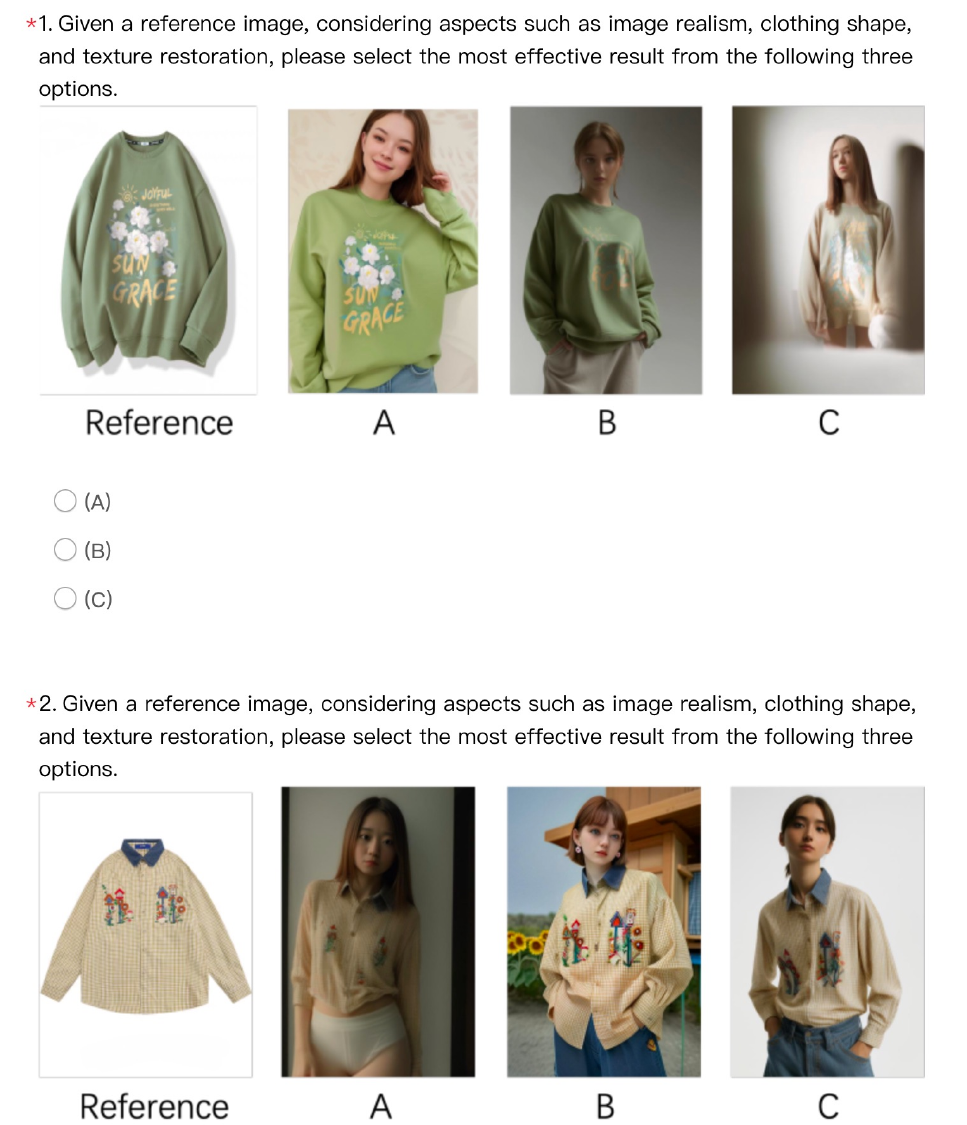}
  \caption{interface of the questionnaire used to evaluate the impact of text prompts.}
  \label{fig:human_sd}
\end{figure*}

\end{document}